\documentclass{article}
\usepackage{kr}

\usepackage{times}
\usepackage{soul}
\usepackage{url}
\usepackage[hidelinks]{hyperref}
\usepackage[utf8]{inputenc}
\usepackage[small]{caption}
\usepackage{graphicx}
\usepackage{amsmath,amssymb}
\usepackage{amsthm}
\usepackage{listings}
\usepackage{subcaption} 

\usepackage{pifont}
\newcommand{\cmark}{\ding{51}}%
\newcommand{\xmark}{\ding{55}}%

\usepackage[dvipsnames]{xcolor}
\usepackage{pifont}
\newcommand{\CMARK}{\textcolor{ForestGreen}{\cmark}}
\newcommand{\XMARK}{\textcolor{BrickRed}{\xmark}}

\usepackage{booktabs}

\usepackage{multicol}
\usepackage{algorithm}
\usepackage{algpseudocode}

\usepackage[dvipsnames]{xcolor}

\newcommand{\cc}[1]{\textcolor{orange}{#1}}

\newcommand{\eg}{\emph{e.g.}~} 

\newcommand{\ie}{\emph{i.e.}~} 

\newcommand{\cf}{\emph{cf.}~}

\usepackage{natbib}

\title{EXPIL: Explanatory Predicate Invention for Learning in Games}

\author{%
Jingyuan Sha$^1$\and
Hikaru Shindo$^1$\and
Quentin Delfosse $^{1}$\and\\
Kristian Kersting$^{1,2,3}$\and
Devendra Singh Dhami$^{4}$\\
\affiliations
$^1$Technical University of Darmstadt\\
$^2$Hessian Center for Artificial Intelligence(hessian.AI)\\
$^3$German Research Centre for Artificial Intelligence (DFKI)\\
$^4$Eindhoven University of Technology\\
\emails
\{jingyuan.sha, hikaru.shindo, quentin.delfosse, kersting\}@cs.tu-darmstadt.de,
d.s.dhami@tue.nl
}

\begin{document}

\maketitle

\begin{abstract}
Reinforcement learning (RL) has proven to be a powerful tool for training agents that excel in various games. However, the black-box nature of neural network models often hinders our ability to understand the reasoning behind the agent's actions. Recent research has attempted to address this issue by using the guidance of pretrained neural agents to encode logic-based policies, allowing for interpretable decisions. A drawback of such approaches is the requirement of large amounts of predefined background knowledge in the form of predicates, limiting its applicability and scalability. In this work, we propose a novel approach, \emph{Explanatory Predicate Invention for Learning in Games (EXPIL)}, that 
identifies and extracts predicates from a pretrained neural agent, later used in the logic-based agents, reducing the dependency on predefined background knowledge. 
Our experimental evaluation on various games demonstrate the effectiveness of EXPIL in achieving explainable behavior in logic agents while requiring less background knowledge.
\end{abstract}

\section{Introduction}
Deep reinforcement learning (RL) agents have revolutionized the field by employing neural networks to make decisions based on unstructured input state spaces, thereby removing the necessity for manual feature engineering~\cite{Mnih2015dqn,alphago,Sutton2018}.
This advancement allows these agents to autonomously learn complex tasks. However, despite their impressive capabilities, these black-box policies present significant challenges. 
One major issue is their lack of \emph{interpretability}~\cite{rudin2019explaining}, which refers to their inability to provide a clear and understandable explanation of the reasoning behind their action selections. 
This opaqueness makes it difficult for humans to trust and verify the decision-making processes of these agents. 
Furthermore, these policies often exhibit a lack of robustness when faced with small environmental changes~\cite{Pinto17robust_icml,wulfmeier2017robust}. 
This fragility can lead to suboptimal policies or even catastrophic performance drops when the agents encounter situations that differ from their training conditions~\cite{Langosco2022goal,delfosse2024hackatari,kohler2024interpretable}, and their inherent lack of interpretability prevents human experts from identifying and correcting potentially misaligned behaviors~\cite{delfosse2024interpretable}.

To address this limitation, neuro-symbolic RL (NeSy-RL) combines the learning power of neural networks with the interpretable nature of symbolic logic and reasoning. 
NeSy-RL policies are not only transparent, but can enhance performances, surpassing purely neural agents and improving generalization capability~\cite{JiangL19NLRL,DelfosseSDK23nudge}.
Their policies encode a weighted set of action rules (\eg distilled from a pretrained neural policy). 
These rules combine a set of state-facts (such as \textit{the agent is close to the enemy} or \textit{the agent above from the treasure}) by applying predefined concepts (\eg \textit{closeby}, \textit{above}) to the detected entities. 
At inference, these state-facts are evaluated in each weighted rule to select an action.

\begin{figure}
    \centering
    \includegraphics[width=\linewidth]{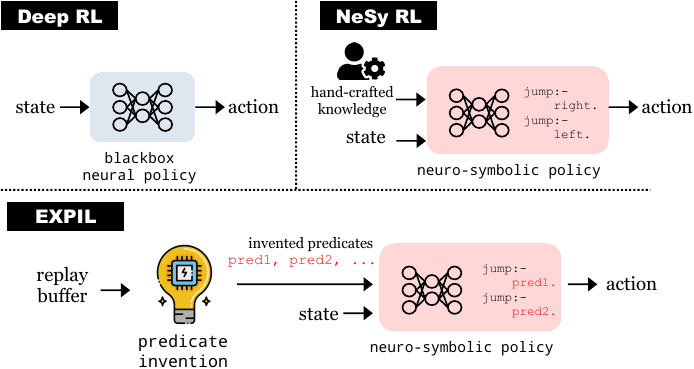}
    \caption{\textbf{EXPIL introduces predicate invention within neuro-symbolic RL agents.} 
    EXPIL extracts concepts from a replay buffer, lated employed to compute optimal actions through neuro-symbolic policies. 
    In contrast to neural policies, EXPIL generates highly interpretable policies using logic and requires few hand-crafted priors compared to conventional neuro-symbolic policies.}
    \label{fig:intro_overview}
\end{figure}

The major flaw of such existing NeSy-RL systems is their reliance on complex, hand-crafted priors (in the form of concepts) provided to the reasoning agents.
Logic based agents need pre-defined state evaluation functions to evaluate interpretable concepts.
Consequently, this severely limits the applicability of the NeSy-RL systems across various domains, in contrast to deep RL agents that require a minimal prior knowledge and achieve high performance by learning from data~\cite{muzero19}. 
This raises the question: \emph{How can we develop NeSy-RL agents that can autonomously discover new concepts while learning how to solve tasks?}


To mitigate this issue, we propose \underline{EX}planatory \underline{P}redicate \underline{I}nvention for \underline{L}earning (EXPIL, \cf~Figure~\ref{fig:intro_overview}).
EXPIL integrates \emph{Predicate Invention (PI)}, a set of techniques to automatically discover new predicates for logic-based machine learning. 
By inventing new predicates from available data, PI significantly reduces the amount of required priors~\cite{pi1988,Kok2007StatisticalPI}.
To perform PI in RL, EXPIL identifies new predicates using general physical concepts and demonstrations of a trained neural agent. These predicates can then be integrated first-order logic policies, thus producing interpretable neuro-symbolic policies without requiring hand-crafted knowledge.



%

Overall, we make the following contributions: 
\begin{itemize}
    \item We propose EXPIL\footnote{\url{https://github.com/ml-research/EXPIL}}, a NeSy-RL framework incorporating predicate invention,
    that produces interpretable policies (compared to deep ones), requiring little background knowledge (compared to conventional NeSy-RL ones).
    \item We empirically show that EXPIL outperforms both purely neural and state-of-the-art NeSy agents in logically challenging RL environments.
    \item We propose two predicate evaluation metrics -- \emph{Necessity} and \emph{Sufficiency} to quantify the probability that a particular observed fact leads to a choice of action. 
\end{itemize}

Hereafter, we introduce the necessary background on Logic, RL and PI for understanding EXPIL.

\section{Background}
Before delving into the EXPIL pipeline, let us establish the formal background of the framework.

\subsection{First-Order Logic (FOL)}
In FOL, a 
{\it Language} $\mathcal{L}$ is a tuple $(\mathcal{P}, \mathcal{D}, \mathcal{F}, \mathcal{V})$,
where $\mathcal{P}$ is a set of predicates, 
$\mathcal{D}$ a set of constants,
$\mathcal{F}$ a set of function symbols (functors), and 
$\mathcal{V}$ a set of variables.
A {\it term} is either a constant (\eg $\texttt{obj1}, \texttt{agent}$), a variable (\eg $\texttt{O1}$), or a term which consists of a function symbol.
An {\it atom} is a formula ${\tt p(t_1, \ldots, t_n) }$, where ${\tt p}$ is a predicate symbol (\eg $\texttt{closeby}$) and ${\tt t_1, \ldots, t_n}$ are terms.
A {\it ground atom} or simply a {\it fact} is an atom with no variables (\eg $\texttt{closeby(obj1,obj2)}$).
A {\it literal} is an atom ($A$) or its negation ($\lnot A$).
A {\it clause} is a finite disjunction ($\lor$) of literals. 
A {\it ground clause} is a clause with no variables.
A {\it definite clause} is a clause with exactly one positive literal.
If  $A, B_1, \ldots, B_n$ are atoms, then $ A \lor \lnot B_1 \lor \ldots \lor \lnot B_n$ is a definite clause.
We write definite clauses in the form of $A~\mbox{:-}~B_1,\ldots,B_n$.
Atom $A$ is called the {\it head}, and set of negative atoms $\{B_1, \ldots, B_n\}$ is called the {\it body}.
We sometimes refer to clauses as \emph{rules}.
\emph{Differentiable Forward Reasoning} is a data-driven approach of reasoning in FOL~\cite{Russel09}. 
In forward reasoning, given evaluated facts and rules, new facts are deduced by applying the facts to the rules.
Differentiable forward reasoning~\cite{Evans18,Shindo2023alphailp} is a differentiable implementation of forward reasoning using tensor operations.

\subsection{Reinforcement Learning (RL)}
In RL, the task is modelled as a Markov decision process, $\mathcal{M}\!= <\!\!\mathcal{S}, \mathcal{A}, P, R\!\!>$, where, at every timestep $t$, an agent in a state $s_t \in \mathcal{S}$, takes action $a_t \in\!\mathcal{A}$, receives a reward $r_t = R(s_t, a_t)$ and a transition to the next state $s_{t+1}$, according to environment dynamics $P(s_{t+1}|s_t,a_t)$. 
Deep agents attempt to learn a parametric policy, $\pi_\theta (a_t|s_t)$, to maximize the return (\ie $\sum_{t} \gamma^t r_t$, with $\gamma \in [0, 1]$). 
However, The desired input-to-output distribution (\ie state to action) is not directly accessible, as RL agents only observe returns. 
The value $V_{\pi_\theta}(s_t)$ (resp. Q-value $Q_{\pi_\theta}(s_t, a_t)$) function provides the expected return of the state (resp. state/action pair) following the policy $\pi_\theta$. 
Policy-based methods directly optimize $\pi_\theta$ using the noisy return signal, which can lead to potentially unstable learning. 
Value-based methods learn to approximate value functions $\hat{V}_\phi$ or $\hat{Q}_\phi$, implicitly encoding the policy (\eg by selecting actions with the highest Q-value with high probability) \cite{Mnih2015dqn}. 





\subsection{FOL for RL}
Following~\cite{DelfosseSDK23nudge}, FOL policies can be created to solve RL challenges. To do so, the set of predicates $\mathcal{P}$ can be divided into a set of action predicates and $\mathcal{P}_A$, and a set of state predicates $\mathcal{P}_S$. 
These are used to form \textit{action rules}. Let $X_A$ be an action atom and $X_S^{(1)}, \ldots, X_S^{(n)}$ be state atoms.
An action rule is a rule, written as $X_A\ \texttt{:-} \ \ X_S^{(1)}, \ldots, X_S^{(n)}$. 
For instance, the action atom could correspond to the environment action \textit{jump}, while the state ones could encode \textit{the agent is close to the enemy}.
The policy is then encoded as a set of weighted action rules. At inference time, each action atom is evaluated using forward reasoning on the facts, which provides an environmental action probability for each action.

\subsection{Predicate Invention}
\emph{Predicate invention (PI)} systems find new predicates to describe or analyze various aspects of a subject or domain. Thereby expanding the system's language and reducing reliance on human experts~\cite{Stahl93PI,aAthakravi12PI}.
\textbf{NeSy-$\boldsymbol{\pi}$}~\cite{sha24} is a neuro-symbolic system that integrates predicate invention with differentiable rule learners~\cite{Shindo2023alphailp} to discover useful relations from complex visual scenes. 
NeSy-$\pi$ invents new predicates to describe observations better using only primitive concepts which are then utilized by the learner to solve classification tasks on complex visual scenes.

\begin{figure*}[th!]
\vskip 0.2in
\begin{center}
\includegraphics[width=0.94\textwidth]{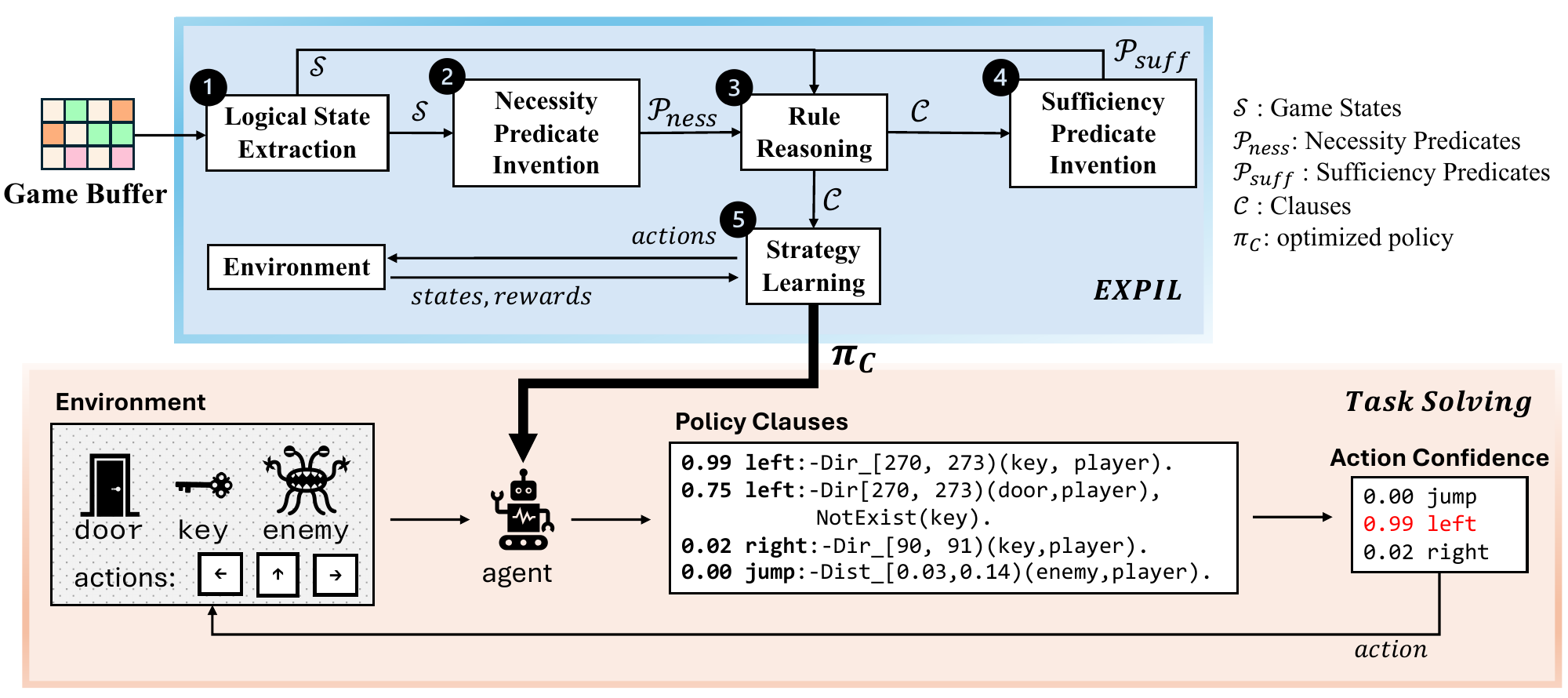}
\caption{
\textbf{EXPIL Architecture}. \textbf{Top:} EXPIL uses a state/action game buffer of pretrained agents to extract logical states $\mathcal{S}$, invents necessity predicates $\mathcal{P}_\mathit{ness}$ and sufficiency predicates $\mathcal{P}_\mathit{suff}$, deduces policy clauses $\mathcal{C}$, and finally learns an optimized policy $\pi_\mathcal{C}$ through interaction with the game environment.
\textbf{Bottom:}
At inference time, the logic agent uses the optimized policy and game states as input, evaluates valid policy clauses from the environment, and selects the action corresponding to the rule with the highest evaluation.
}
\label{fig:archi}
\end{center}
\vskip -0.2in
\end{figure*}

\section{EXPIL}
EXPIL produces logic agents capable of deducing weighted policy clauses $\mathcal{C}_w$ from a replay buffer $\mathcal{B}$, using limited background knowledge. 
Its architecture, depicted in Figure~\ref{fig:archi}, consists of five major components:\\
(1) \textbf{Logical State Extraction.} The object-centric game states-action pairs $\mathcal{S}$ are extracted from the provided replay game buffer, $\mathcal{B}$, from pretrained agents. \\
(2) \textbf{Necessity Predicate Invention.} Necessity predicates $\mathcal{P}_\mathit{ness}$ are invented using $\mathcal{B}$ to capture essential properties or relationships for certain actions in the game.\\
(3) \textbf{Rule Reasoning.} The invented predicates are utilized to infer policy clauses $\mathcal{C}_w$ that provide better explanations for the behaviors observed in the replay buffer. 
Simultaneously, the rule scores $\mu$ are evaluated using the game buffer.\\
(4) \textbf{Sufficiency Predicate Invention.} Based on the reasoned policy clauses, sufficiency predicates $\mathcal{P}_\mathit{suff}$ are invented with the aim of enriching predicate expressiveness. 
These predicates capture additional conditions or factors that contribute to the sufficiency of certain actions.\\
(5) \textbf{Strategy Learning}. Through gameplay, the weights $w$ of the policy clauses $\mathcal{C}_w$ are learned to improve the performance of the agent.\\
Let us now describe each step in detail. 

\subsection{Logical State Extraction}
\label{sec:neural_guided_symbolic_abstraction}
To perform PI, we first generate a labeled dataset, corresponding to a game buffer,
of human experts or pretrained neural agents rollouts.
As object-detection is not our focus, we do not use object discovery techniques~\cite{delfosse2023boosting,luo2024insight} to train logic agents, we directly make use of object-centric representations, similar to ones from~\cite{ocatari}. Every state directly consists of a set of objects with their attributes. 
For example, an enemy in a game state can be represented by its position and a distinctive identifier.



Let $\mathcal{S}$ denote the game buffer, \ie the set of pairs of a state and an action,
and $\mathcal{A}$ denote the action space of the game. 
For each action $a\in\mathcal{A}$, we decompose $\mathcal{S}$ to positive and negative sets:
$\mathcal{S}_a^+ = \{ (s, a^\prime) ~|~ (s, a^\prime) \in \mathcal{S} \land a^\prime = a\}$
and
$\mathcal{S}_a^- = \{ (s, a^\prime) ~|~ (s, a^\prime) \in \mathcal{S} \land a^\prime \neq a\}$.
Throughout the paper, we refer to $\mathcal{S}_a^+$ as a set of positive states of action $a$, and $\mathcal{S}_a^-$ as a set of negative states of action $a$.

\subsection{Necessity Predicate Invention}
\label{sec:necessity_predicate_invention}
EXPIL discovers new predicates automatically by using pre-defined physical concepts. 
A \textit{pre-defined physical concept} refers to a pre-defined function that maps pairs of objects to specific values or ranges. 
EXPIL uses two pre-defined physical concepts: \emph{distance} and \emph{direction}. 
The \textit{distance} calculates the space between two objects, while the \textit{direction} determines the angle of one object relative to another.

To discover useful task-specific concepts using the pre-defined physical concepts, we consider parameterized predicates representing various degrees of distance and direction.
For this, we introduce \textit{reference range}, which is defined as a valid range of values of distance and direction.
If the relation or property of the objects locates lies within the reference range, the predicate is evaluated as \textit{true}, otherwise, it is evaluated as \textit{false}.
For example, consider a reference range $\texttt{[0,1)}$ associated with the concept $\texttt{distance}$. In this case, a predicate $\texttt{Distance\_[0,1)}$ can be interpreted as a function to determine \textit{whether the distance between two objects is between 0 and 1 (inclusive of 0 but exclusive of 1)}.

Using the reference range, EXPIL first generates candidates of predicate and selects only promising ones by evaluating them with a heuristic.
To cover various degrees of distance and directions, EXPIL generates new predicates efficiently by increasing intervals of reference range uniformly. For example, suppose we want to specify the distance from $1$ meter to $100$.
This can be achieved by considering the truth value of one of the following predicates:
\begin{lstlisting}
Predicate 0:  Distance_[0,1)
Predicate 1: Distance_[1,2)
...
Predicate 99: Distance_[99,100)
\end{lstlisting}
\eg $\texttt{Distance\_[0,1)(agent,enemy)}$ represents the fact that the agent and the enemy are distant of less than $1$ meter \ie they are very close to each other.

\begin{figure*}[t!]
\vskip 0.2in
\begin{center}
\includegraphics[width=0.9\textwidth]{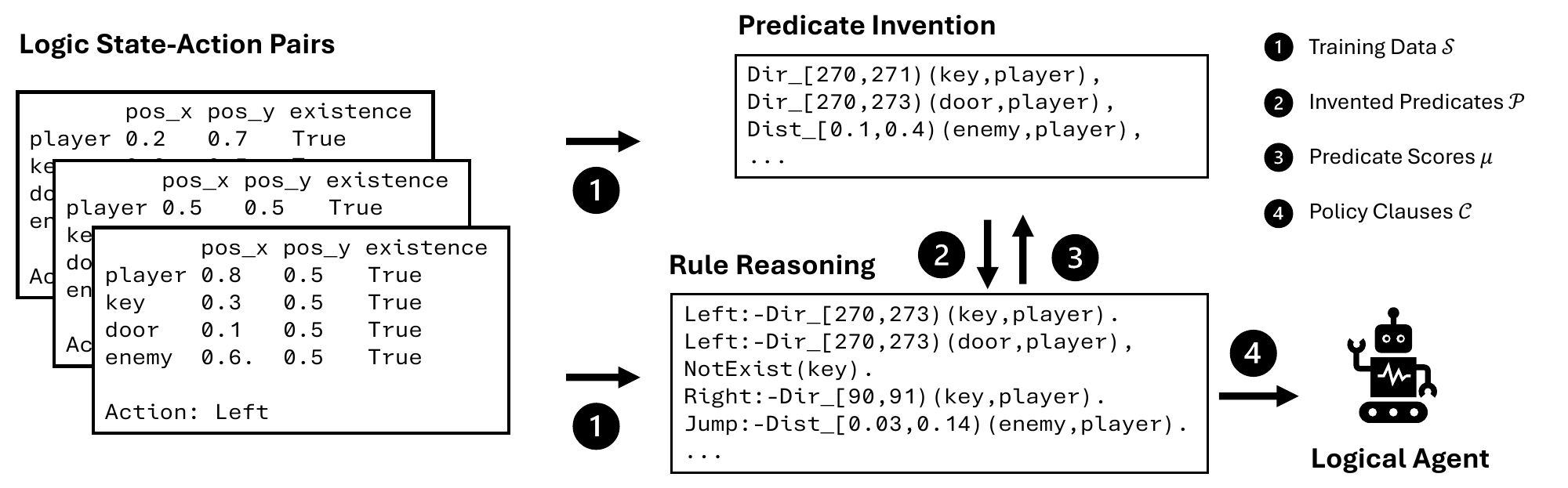}
\caption{
\textbf{The Predicate Invention module of EXPIL}. EXPIL utilizes an object-centric state-action buffer as input for predicate invention and rule reasoning. 
Candidate predicates are invented to serve and combined within clauses of the policy. 
The reasoned policy clauses are evaluated, and promising rules are selected as output to the logical agent.
}
\label{fig:pi_main}
\end{center}
\vskip -0.2in
\end{figure*}

In practice, most of the newly generated predicates are not critical for solving a specific task. For example, the predicate $\texttt{Distance\_[0,1)}$, \ie the concept of being closeby, would be more important than the predicate $\texttt{Distance\_[99,100)}$, \ie the concept of being distant of one specific long distance, since the former can contribute to effective action taking to survive longer (\eg avoiding enemies). 
We consider such important predicates as \emph{necessary} predicates to solve the environment. In this steps, EXPIL invents such predicates.
We follow the NeSy-$\pi$~\cite{sha24} approach to evaluate the \textit{necessity} of each predicate. 
Let us now devise a formal definition.
\paragraph{Necessity:} The necessity of a logical expression $e$ on action $a$, denoted as $\mu_e(\mathcal{S}_a^+)$, measures the cumulative confidence of the logical expression across all states in $\mathcal{S}_a^+$. 
Higher necessity values indicate that more states of $\mathcal{S}_a^+$ evaluate to $true$ for $e$. 
\begin{equation}
 \label{eq:ness}
    \mu_e(\mathcal{S}_a^+) = \frac{1}{|\mathcal{S}_a^+|} \sum_{s\in\mathcal{S}_a^+} r_e(s)
\end{equation}
where $\mathcal{S}_a^+$ represents the buffered game states resulting from taking action $a$, $r_e(s)$ is a differentiable forward reasoning function \cite{Shindo2023alphailp} with respect to logical expression, which returns the confidence of the evaluation result of the expression $p$ for state $s$. In this paper, the logical expression includes predicates and policy clauses.


For example, in the environment depicted in Figure~\ref{fig:archi}, let us consider an action jump ($\uparrow$), and let $\mathcal{S}^+_{\uparrow}$ be a set of positive states in the game buffer where a pretrained agent selected the action to jump $\uparrow$.
Well-trained agents would jump to avoid enemies when they get close to the agent, and thus,  the concept of $\emph{closeby}$ would often be observed in the positive states $\mathcal{S}^+_{\uparrow}$. To this end, predicates that are relevant to the concept of \emph{closeby} (\eg  $\texttt{Distance\_[0,1)}$) would get high evaluation scores by Eq.~\ref{eq:ness}.

\subsection{Rule Reasoning}
\label{sec:rule_reasoning}
Using the invented necessary predicates, EXPIL searches a set of promising action rules by performing beam search.
Intuitively, EXPIL extensively generates new candidates of action rules using the invented necessary predicates and evaluates them with heuristics to select only promising ones. 
EXPIL iterates this rule generation until it gets sufficiently complex action rules to solve the environment.

The necessity predicates $\mathcal{P}_\mathit{ness}$ are invented to evaluate the facts within the game states and are utilized as fundamental components of the policy clauses $\mathcal{C}$. Let $\mathcal{L}$ represent a language containing all the invented predicates $\mathcal{P
}$. By combining multiple predicates $P_1(X), P_2(X),..., P_n(X)$ as the antecedents and the action $A(X)$ as the consequent, a game rule $C$ can be constructed as follows:

\begin{lstlisting}
C: A(X):-P1(X),P2(X),...,Pn(X).
\end{lstlisting}
The rules are searched action by action. They are interpreted as \textit{if the antecedents $P_1(X)$, $P_2(X)$, ..., $P_n(X)$ are true, then take the action $a$}.
For each action $a$, rules are generated incrementally and stored in a rule set $\mathcal{C}_a$. 
These rules are searched in a top-down manner. Initially, the rule set $\mathcal{C}_a$ contains only one initial rule with no atoms in its body, which evaluates to $true$ for any state. 
Subsequently, in each step, each rule in $\mathcal{C}_a$ is extended with each of the atoms in the language $\mathcal{L}$. 

For example, the initial rule set for the action $\texttt{Left}$ is represented as $\mathcal{C}_0 = \{C_0\}$, with
\begin{lstlisting}
C_0 Left(X):-.
\end{lstlisting}
which is interpreted as \textit{take action left if true}. 
If $\mathcal{L}$ contains atoms $\texttt{Dir\_[0,1)(enemy, player)}$ and $\texttt{Dir\_[1,2)(enemy, player)}$, the first step of extension generates a new rule set $\mathcal{C}_1 =\{C_1, C_2\}$
\begin{lstlisting}
C_1 Left(X):-Dir_[0,1)(enemy,player).
C_2 Left(X):-Dir_[1,2)(enemy,player).
\end{lstlisting}
Similarly, the second step of extension further extends each rule from $\mathcal{C}_1$ using each of the atoms in the language $\mathcal{L}$. This process is repeated for $N$ steps to collect rules with bodies of varying lengths.
 
Each reasoned rules $C\in\mathcal{C}_i, 1\le i\le N$, is then evaluated for its necessity $\mu_{C}(\mathcal{S}_a^+)$ and sufficiency $\mu_{C}(\mathcal{S}_a^-)$ (\cf Section~\ref{sec:suff_inv}). These evaluations determine how necessary and sufficient the rules are for the given actions. The rules are then ranked based on their necessity, and the Top-$k$ rules are selected for strategy learning as described in Section~\ref{sec:strategy_learning}.

\subsection{Sufficiency Predicate Invention}
\label{sec:suff_inv}
To enhance the performance of logic-based agents, EXPIL combines invented necessary predicates to compose more expressive predicates. 
This can be achieved by computing the disjunction of rules produced by the rule reasoning step, as described in Sec.~\ref{sec:rule_reasoning}. 
For each action $a\in \mathcal{A}$, EXPIL aims to invent predicates that are less correlated to other actions $a^\prime$ $(a^\prime \in \mathcal{A}, a^\prime \neq a)$.
We call the resulted predicates \emph{sufficiency predicates}, as such predicates are motivated by searching for a sufficient condition of the action.


Given a set of rules $\mathcal{C}_a$ for an action $a$. Any subset $\mathcal{C}_{p} \subset \mathcal{C}_a$, such that $2\le |\mathcal{C}_p|\le |\mathcal{C}_a|$ defines a new predicate $p$. This subset, $\mathcal{C}_p$, is referred to as the \textit{explanation clause set} of the predicate $p$. The meaning of $p$ is interpreted as the disjunction of the clauses within $\mathcal{C}_p$. 

For example, consider the rules set $\mathcal{C}=\{C_1, C_2, C_3\}$:
\begin{lstlisting}
C1: Jump(X):-Dist_[0,1)(enemy,player,X).
C2: Jump(X):-Dist_[1,2)(enemy,player,X).
C3: Jump(X):-Dist_[2,3)(enemy,player,X).
\end{lstlisting}
A sufficiency predicate $\texttt{SuffPred}$ can be invented by taking the disjunction of these three clauses, i.e. $\texttt{SuffPred} = C_1 \lor C_2 \lor C_3$.
\begin{lstlisting}
SuffPred(X):-Dist_[0,1)(enemy,player,X).
SuffPred(X):-Dist_[1,2)(enemy,player,X).
SuffPred(X):-Dist_[2,3)(enemy,player,X).
\end{lstlisting}
The invented predicate, $\texttt{SuffPred}$, interprets the concepts that \textit{if the distance between the enemy and the player is in the range of $0$ to $3$}. Each clause in $\mathcal{C}_p$ represents a \textit{variation} of the invented predicate.
By taking the disjunction of multiple clauses, the invented predicate can expand its range to cover more states (\eg measuring distances from $0$ to $3$ by taking three individual predicates) and thereby becomes much more expressive. 

Theoretically, there are $2^{|\mathcal{C}_a|-1-|\mathcal{C}_a|}$ predicates that can be invented from a clause set $\mathcal{C}_a$, ensuring that any sufficient predicate includes at least two clauses. Due to this exponential growth rate, it is crucial to evaluate the invented predicates appropriately to select the most useful ones. 

EXPIL evaluates these predicates considering their sufficiency. The \textit{sufficiency} of a logical expression, as proposed by \cite{sha24}, is defined as follows.

\paragraph{Sufficiency:} The sufficiency of a logical expression $e$, denoted as $\mu_e(\mathcal{S}_a^-)$,  measure the inverse of the cumulative confidence of $e$ in all the states in $\mathcal{S}_a^-$. The higher the \textit{sufficiency} of $e$, the fewer states in $\mathcal{S}_a^-$ for which it holds true.
\begin{equation}
\label{eq:suff}
    \mu_e(\mathcal{S}_a^-) = \frac{1}{|\mathcal{S}_a^-|} \sum_{s\in\mathcal{S}_a^-} (1-r_e(s))
\end{equation}
For a clause set $\mathcal{C}_a$, EXPIL clusters the set into multiple clusters based on the relations and the objects involved. Only the rules with same objects and same relations but different reference ranges are combined. For example, if $C_4$ and $C_5$ are two rules in $\mathcal{C}_a$ as follows:
\begin{lstlisting}
C4(X):-Dist_[0,1)(enemy,player,X).
C5(X):-Dist_[1,2)(key,player,X).
\end{lstlisting}
they are not combined as the same cluster since they have different objects in the predicate.

The necessity scores of disjunctive clusters are typically high because they combine the reference ranges. For example, a predicate can be invented by clustering all the clauses that evaluate the distance between the player and the enemy, which can be written as $\texttt{Dist\_[0,100)(enemy, player)}$. If $100$ is the maximum distance in the game and the player and the enemy exist in all states, this predicate can always be \textit{true}, thus reaching a necessity score of $1$ according to Equation \ref{eq:ness}. However, its sufficiency score would be $0$ according to Equation \ref{eq:suff} because it is also true in all negative states. 

By inventing sufficiency predicates from disjunction, EXPIL first clusters the clauses with the same objects and relations, but different reference ranges, resulting in high necessity for the clusters. Then, to increase their sufficiency, clauses are gradually removed from the cluster until the sufficiency of the cluster reaches a given threshold. 
The clause removed in each step is chosen as the least sufficient one (\textit{i.e.} the clause with removing it can improve the efficiency most) to ensure the remaining clauses provide the best possible increment in sufficiency. As clauses are removed from the cluster, its necessity will also decrease; however, as long as it remains above zero, it is valuable for the rule reasoning. By refining the predicates through disjunction and selective reduction, the system effectively balances the necessity and sufficiency to enhance the expressiveness and utility of the predicates in rule reasoning.

\begin{figure}[t]
\vskip 0.2in
\begin{center}
\includegraphics[width=\columnwidth]{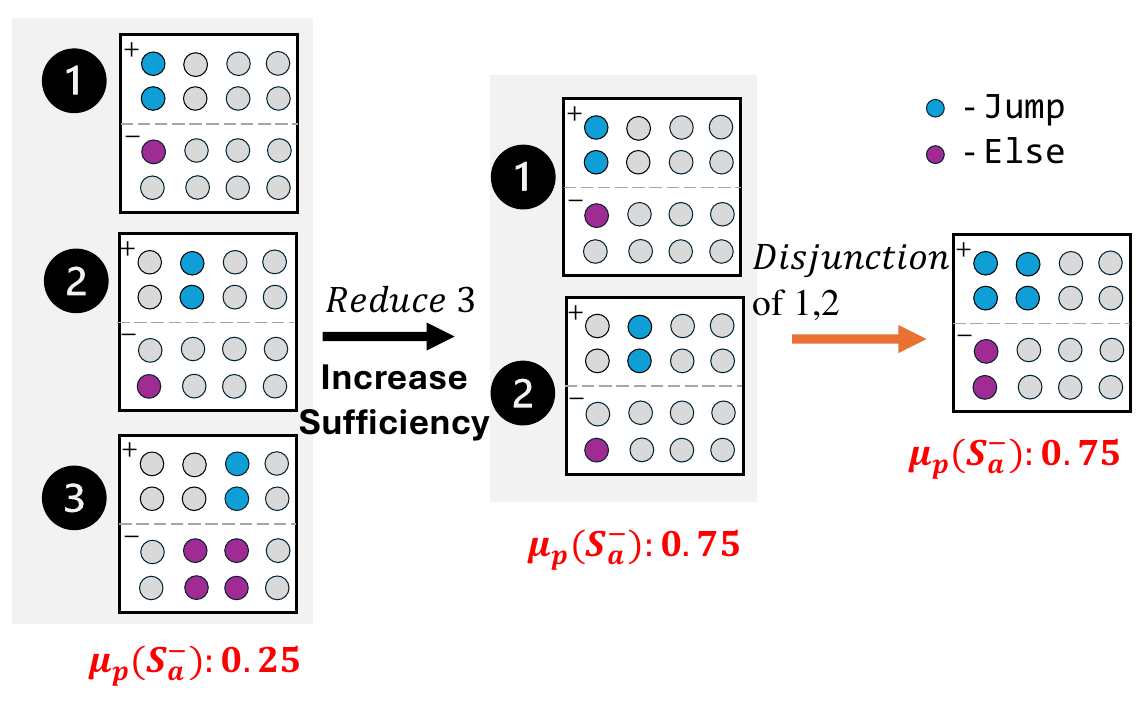}
\caption{An example step for sufficiency predicate invention. 
In each rectangle box: the $8$ circles above the dashed line indicate the states taking the action $\texttt{Jump}$, while the remaining circles indicate the states taking other actions.
Boxes $1, 2, 3$ shows the evaluation results of three different predicates, with blue circles of box $i\in[1,2,3]$ representing the positive states that satisfy the predicate $i$ and purple circles of box $i\in[1,2,3]$ representing the negative states that satisfy the predicate $i$. The $\mu_p(\mathcal{S}_a^+)$ and $\mu_p(\mathcal{S}_a^-)$ at the bottom indicate the scores of the invented sufficiency predicate.
}
\label{fig:suff_pred}
\end{center}
\vskip -0.2in
\end{figure}

For example, Figure~\ref{fig:suff_pred} illustrates a step in the process of inventing a sufficiency predicate. 
The evaluation result of $3$ different predicates are shown in the left column. 
Each individual predicate has a low necessity but high sufficiency. 
Their disjunction achieves a high necessity but low sufficiency (see the value at the bottom). 
When the third predicate is removed (second column of Figure~\ref{fig:suff_pred}), the disjunction of the remaining two predicates attains a higher sufficiency. 

\subsection{Strategy Learning}
\label{sec:strategy_learning}
After the extraction of a set of good candidate rules to form the policy, a set of randomly initialized weights $W$ is assigned to each policy clause in $\mathcal{C}$. 
These weights are optimized based on an actor-critic method~\cite{KondaT_actor_critic} that back propagates the gradients from the critic to the differentiable logic actors, allowing an update of each weight.
Following~\cite{DelfosseSDK23nudge}, we update both the rule weights and the critic-network weights in each iteration.
To obtain a probabilistic distribution, \textit{softmax} over the evaluation of all actions is taken.

\begin{algorithm}[tb]
\caption{Predicate Invention in EXPIL}
\label{alg:nesy_pi}
\textbf{Input}: $\mathcal{L}_0, \mathcal{S},\mathcal{B}, \mathcal{A}=[\mathtt{left},\mathtt{right},...]$\\
\textbf{Parameter}: $N_\mathit{max\_c\_length},$\\
\textbf{Output}: $\mathcal{L}$
\begin{algorithmic}[1] 
\For {$a \leftarrow \mathcal{A} $}
    \State $\mathcal{S}^+_a,\mathcal{S}^-_a = \mathtt{split\_states}(\mathcal{S}, a) $
    \State \cc{// Nessicity Predicate Invention}
    \State $\mathcal{P}_\mathit{ness} \gets \mathtt{ness\_inv}(\mathcal{L}_0, \mathcal{B})$
    \State $\mu_\mathit{ness} \gets \mathtt{eval}(\mathcal{P}_\mathit{ness}, \mathcal{S}_a^+)$
    \State $\mathcal{P}_\mathit{ness}^* \gets \mathtt{top\_k}(\mathcal{P}_\mathit{ness}, \mu_\mathit{ness})$
    \State $\mathcal{L} \gets \mathtt{update}(\mathcal{L}_0, \mathcal{P}_\mathit{ness}^*)$ 
    \For{ $N_\mathit{step\_max} \leftarrow [1,2,..., N_\mathit{max\_c\_length}]$}
        \State \cc{// Clause Searching}
        \State $\mathcal{C} = \mathtt{init\_clause}(a)$
        \State $\mathcal{C}\_\mathit{step} =$ []
        \For{ $N_\mathit{step} \leftarrow [1,2,...,  N_\mathit{step\_max}] $}
            \State $\mathcal{C} \gets \mathtt{extend}(\mathcal{C}, \mathcal{L})$
            \State $\mu_\mathit{ness},\mu_\mathit{suff} \gets \mathtt{eval\_c}(\mathcal{C}, \mathcal{S}^+_a,\mathcal{S}^-_a)$
            \State $\mathcal{C}\_\mathit{step} \gets \mathtt{top\_k}(\mathcal{C}, \mu_\mathit{ness})$
        \EndFor 
        \State \cc{// Sufficiency Predicate Invention}
        \State $\mathcal{P}_\mathit{suff} \gets \mathtt{suff\_inv}(\mathcal{C}\_\mathit{step}, t_{s})$ 
        \State $\mathcal{P}_\mathit{suff}^* \gets \mathtt{top\_k}(\mathcal{P}_\mathit{suff}, \mu_\mathit{ness})$
        \State $\mathcal{L} \gets \mathtt{update}(\mathcal{L},\mathcal{P}_\mathit{suff}^*)$ 
        
    \EndFor
\EndFor 
\State \textbf{return} $\mathcal{L}$
\end{algorithmic}
\end{algorithm}

\subsection{Algorithm for Predicate Invention in EXPIL}
Algorithm~\ref{alg:nesy_pi} shows the pseudocode of predicate invention process in EXPIL. The algorithm takes as input an initial language $\mathcal{L}_0$, game state-action buffer $\mathcal{S}$, predefined physical concepts $\mathcal{B}$, and the game action space $\mathcal{A} = [\texttt{left}, \texttt{right},...]$, which depends on the specific game environment. It outputs the language $\mathcal{L}$ with invented predicates.
\textbf{(Line 1-2)} For each action $a\in \mathcal{A}$, predicate invention and rule reasoning are performed separately. The replay buffer $\mathcal{S}$ is spit into two groups: $\mathcal{S}^+_a$ contains states with action $a$ and $\mathcal{S}^-_a$ contains states with other actions.
\textbf{(Line 3-7)} Necessity predicates are invented using predefined physical concepts $\mathcal{B}$. The  predicate necessity scores $\mu_\mathit{ness}$ are calculated over the positive states $\mathcal{S}^+_a$. The Top-$k$ predicates $\mathcal{P}^*_\mathit{ness}$ are selected, and the language $\mathcal{L}$ is updated accordingly. 
\textbf{(Line 8-16)} policy clauses are searched starting from the most general clause. For example, to search rules with action $\texttt{left}$, the $\texttt{init\_clause()}$ returns the initial clause set $\mathcal{C}= \{\texttt{left(X):-.}\}$. The clauses are extended with predicates from the language $\mathcal{L}$ in each step, updating the clause set $\mathcal{C}$. Clauses are extended for a maximum of $N_\mathit{step\_max}$ iterations. During the first iteration, the initial clause in $\mathcal{C}$ is extended with new atoms and evaluated against the game buffer $\mathcal{S}$. The Top-$k$ scoring clauses are retained, while the others are pruned. 
\textbf{(Line 17-20)} Sufficiency predicates $\mathcal{P}_\mathit{suff}$ are invented based on the extended clauses in $\mathcal{C}_\mathit{step}$ and a given threshold $t_s$ to ensure the sufficiency is above $t_s$.
The number of sufficiency predicates is limited by retaining only the top $k$. These invented predicates are then added to the language $\mathcal{L}$. 

\section{Experiments}
\begin{figure}[t]
\begin{center}
\begin{subfigure}[b]{0.15\textwidth}
\centering
\includegraphics[width=\columnwidth]{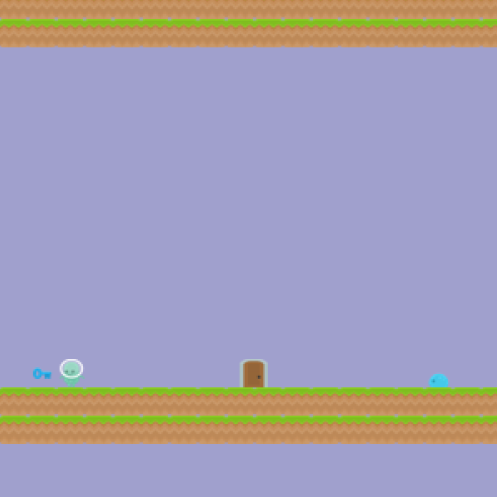}
\caption{Getout}
\label{fig:getout}
\end{subfigure}
\hfill
\begin{subfigure}[b]{0.15\textwidth}
\centering
\includegraphics[width=\columnwidth]{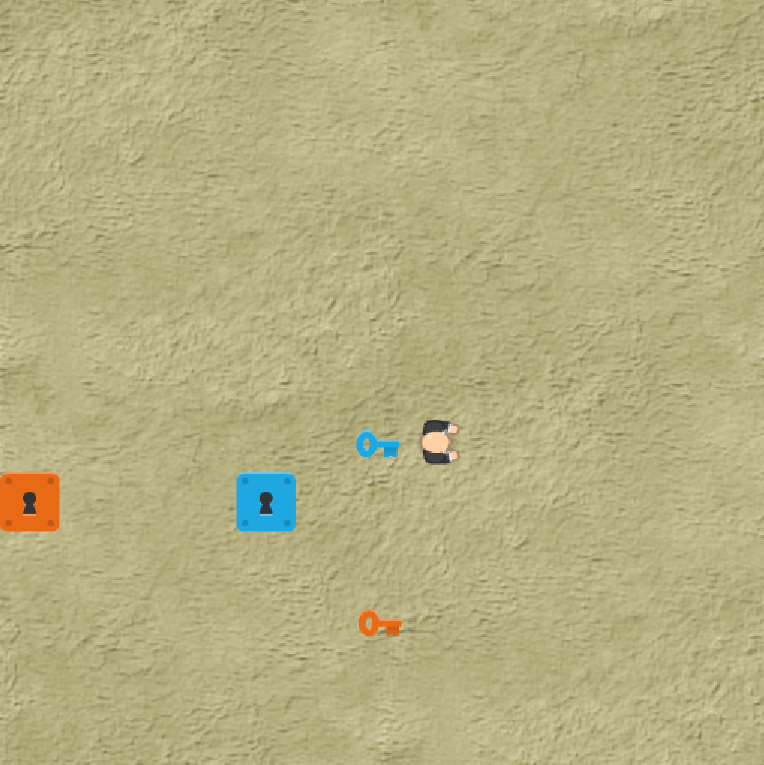}
\caption{Loot}
\label{fig:loot}
\end{subfigure}
\hfill
\begin{subfigure}[b]{0.15\textwidth}
\centering
\includegraphics[width=\columnwidth]{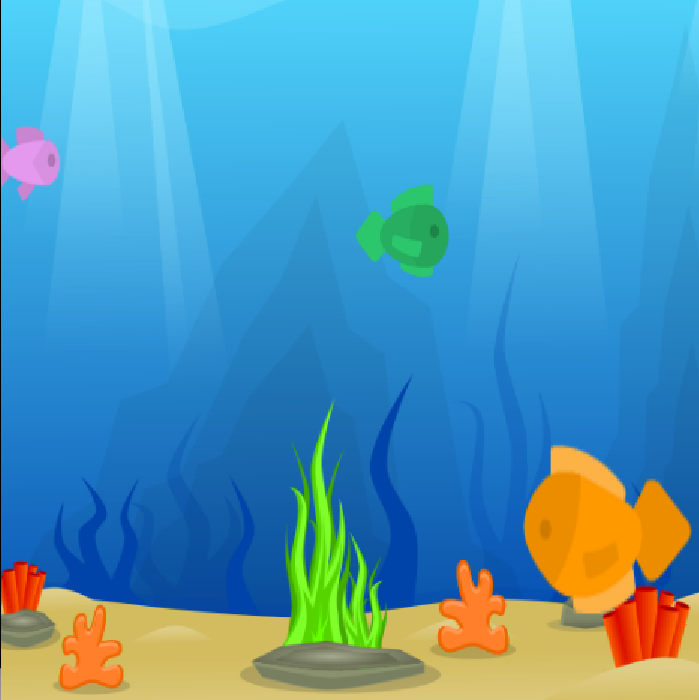}
\caption{Threefish}
\label{fig:threefish}
\end{subfigure}
\caption{Environments used to evaluate EXPIL and baselines.}
\label{fig:game_screenshots}
\end{center}
\end{figure}

To evaluate the performance of predicate invention in the RL setting, we employ \emph{three} established logically challenging environments, which have been used to evaluate state-of-the-art neuro-symbolic RL agents~\cite{DelfosseSDK23nudge}. Exemplary states from each environment are shown in Figure~\ref{fig:game_screenshots}, and let us describe each environment in detail.
\textbf{Getout} is a game that involves the agent moving on a $1.5$D
map with taking the actions $\texttt{left}$, $\texttt{right}$ and $\texttt{jump}$. 
The agent can move freely along $x$-axis but has limited movement along the $y$-axis, which is controlled by the action $\texttt{jump}$. 
The objective is to collect a key on the ground and then go to the door while avoiding an enemy that moves around the map.
In each new epoch, the positions of all objects are randomly placed.
In \textbf{Loot}, the agent moves on a $2$D map, taking actions $\texttt{left}$, $\texttt{right}$, $\texttt{up}$ and $\texttt{down}$. The agent can move freely along both the $x$ and $y$ axes. There are one or two pairs of locks and keys randomly generated in each new game. Locks and keys have IDs, and a key can only open the lock with the corresponding ID. The objective is to collect keys and open corresponding locks until no locks remain. 
In \textbf{Threefish}, the agent moves on a $2$D map, taking actions $\texttt{left}$, $\texttt{right}$, $\texttt{up}$, $\texttt{down}$ and $\texttt{noop}$. The agent can move freely along both the $x$ and $y$ axes. 
The objective is to eat smaller fish while avoiding the bigger fish than the player. 

For baselines, we used the standard neural PPO~\cite{SchulmanWDRK17PPO} and NUDGE~\cite{DelfosseSDK23nudge}, a SOTA NeSy-RL agent.
For a fair comparison, we did not provide the task-specific predicates to the models, and thus they need to acquire them by learning in the environments. 


\textbf{Preliminaries.} 
Two physics concepts are given as background knowledge for predicate invention. \textit{distance} measures the distance between two given objects. 
\textit{direction} measures the direction of one object with respect to another object.
Additionally, the absence of an object is evaluated by the predicate $\texttt{NotExist(O1)}$, which is particularly required for the game \textit{Getout} and \textit{Loot}. The predicate is used to determine whether a specific object is present in a game state. Besides, no further game-specific knowledge is given.

\textbf{Symbolic States Extraction.}
To facilitate predicate invention and rule reasoning, a teacher agent plays the game and collects a \textit{game buffer} containing symbolic game states and corresponding actions. For each game, we randomly select $800$ state-action pairs for each action. Each state records the existence of objects and their positions on the $x$ and $y$ axes. A state is saved in a matrix with shape of $(n+2)\times n$, where $n$ is the number of objects in the game. For the $i$-th object in the state: its existence is saved at position $(i,i)$, its $x$ position is saved at $(i, n+1)$, its $y$ position is saved at $(i, n+2)$. The action for each state is recorded as its index in the action space. This structured data serves as the learning material for the EXPIL system to invent predicates and reason about the policy clauses.

\begin{figure}[h!]
\begin{center}
\includegraphics[width=0.81\columnwidth]{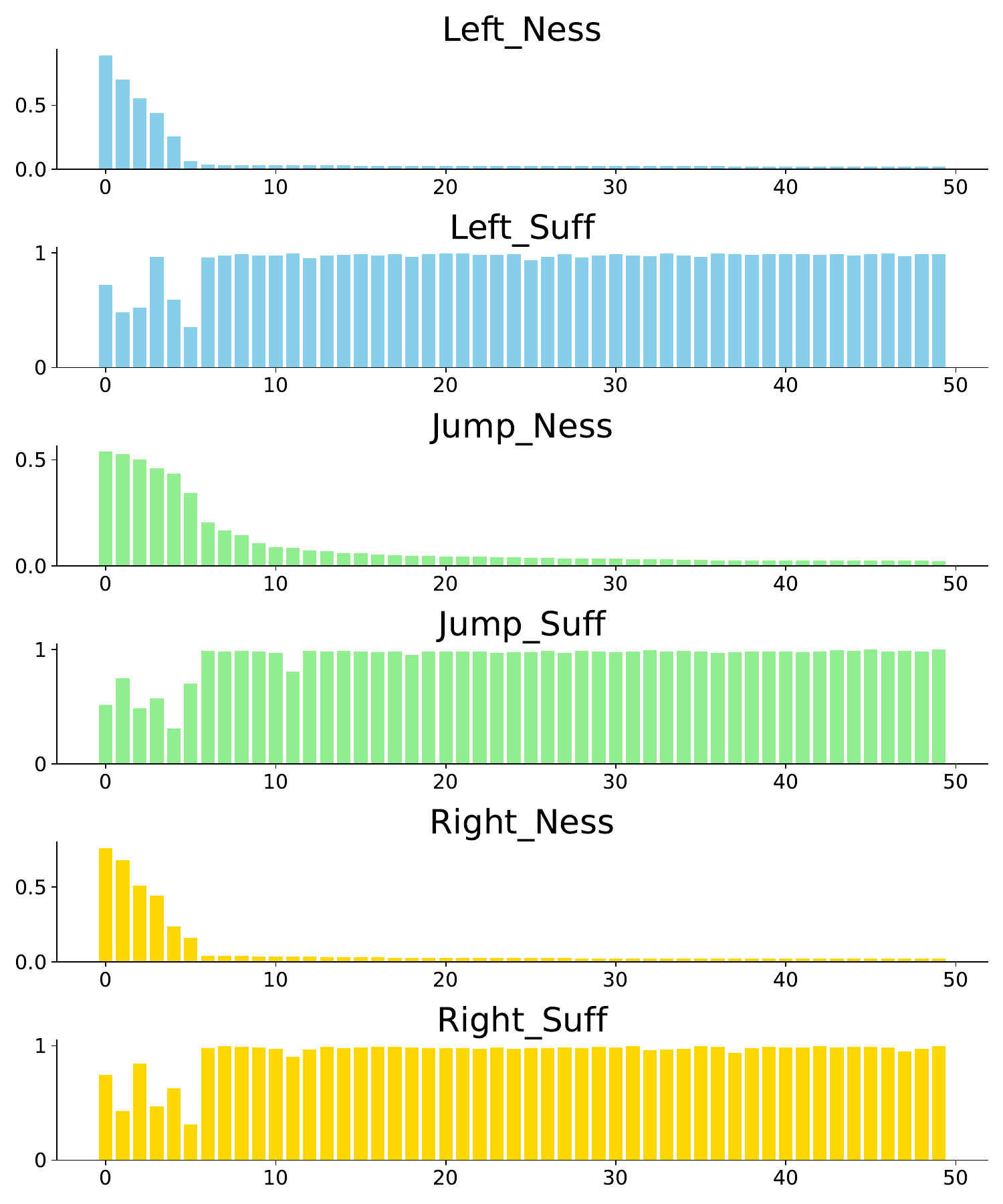}
\caption{\textbf{Necessity Predicate Scores}. The evaluation results of the invented necessity predicates for the game \textit{Getout} (Only the Top-50 predicates ranked by necessity score are displayed). 
\textbf{Red}: scores for the action $\texttt{Left}$; \textbf{Yellow}: scores for the action  $\texttt{Jump}$; \textbf{Green}: scores for the action $\texttt{Right}$. For each action, the necessity scores are ranked in descending order. The corresponding sufficiency scores are aligned with the rank of the necessity scores for the same action.
}
\label{fig:ness_pred_scores}
\end{center}
\end{figure}

\textbf{Necessity Predicate Invention.}
We firstly set out to verify the efficiency of using necessity and sufficiency as metrics for predicate evaluation. 
Figure~\ref{fig:ness_pred_scores} displays the scores of the invented necessity predicates for the game \textit{Getout}. The predicates are invented for each action individually. 
By using an interval unit with $1\%$ of the maximum distance and $4\%$ of the maximum direction. Over $800$ necessity predicates are invented to evaluate different distances and directions between the agent and other three objects (door, enemy, key). The figure only shows the top 50 predicates ranked by necessity scores. 

Predicates with high necessity scores (above 0.1) are selected as promising candidates based on their high necessity and can further be used for the game rule reasoning. 
Although the reference range is chosen to be very small, dividing the distance into $100$ sections and the direction into up to $90$ sections, most of the necessity predicates achieve high sufficiency (close to 1) and low necessity (close to 0) eventually. This indicates that most of the predicates are rarely evaluated as true in both positive and negative states, thus can be pruned to reduce computation complexity.

Table~\ref{tab:ness_percent} shows the percentage of necessity predicates that exceed various thresholds in the game \textit{Getout}. By diving the distance and direction into $100$ and $90$ sections respectively, only $22.3\%$ of the predicates have necessity scores greater than $0.01$, implying that around $78\%$ of the predicates are rarely present in the positive training states.

\begin{table}[th!]
\centering
\begin{tabular}{lrrr}
\toprule
               &  $s>0.001$   &  $s>0.01$  & $s>0.1$ \\
\midrule
$\texttt{left}$  & $100\%$ & $22.3\%$  & $0.9\%$ \\
$\texttt{right}$ &  $100\%$ & $25.1\%$ & $1.0\%$ \\
$\texttt{jump}$  &  $100\%$ & $15.0\% $  &  $1.7\%$\\
\bottomrule
\end{tabular}
\caption{Percentage of the necessity predicates that have necessity score $s$ greater than a given threshold from game \textit{Getout}. The number of reference ranges for distance and direction are $100$ and $90$, respectively. $\uparrow$ necessity threshold = $\downarrow$ remaining predicates.}
\label{tab:ness_percent}
\end{table}

\begin{figure}[th!]
\begin{center}
\includegraphics[width=0.91\columnwidth]{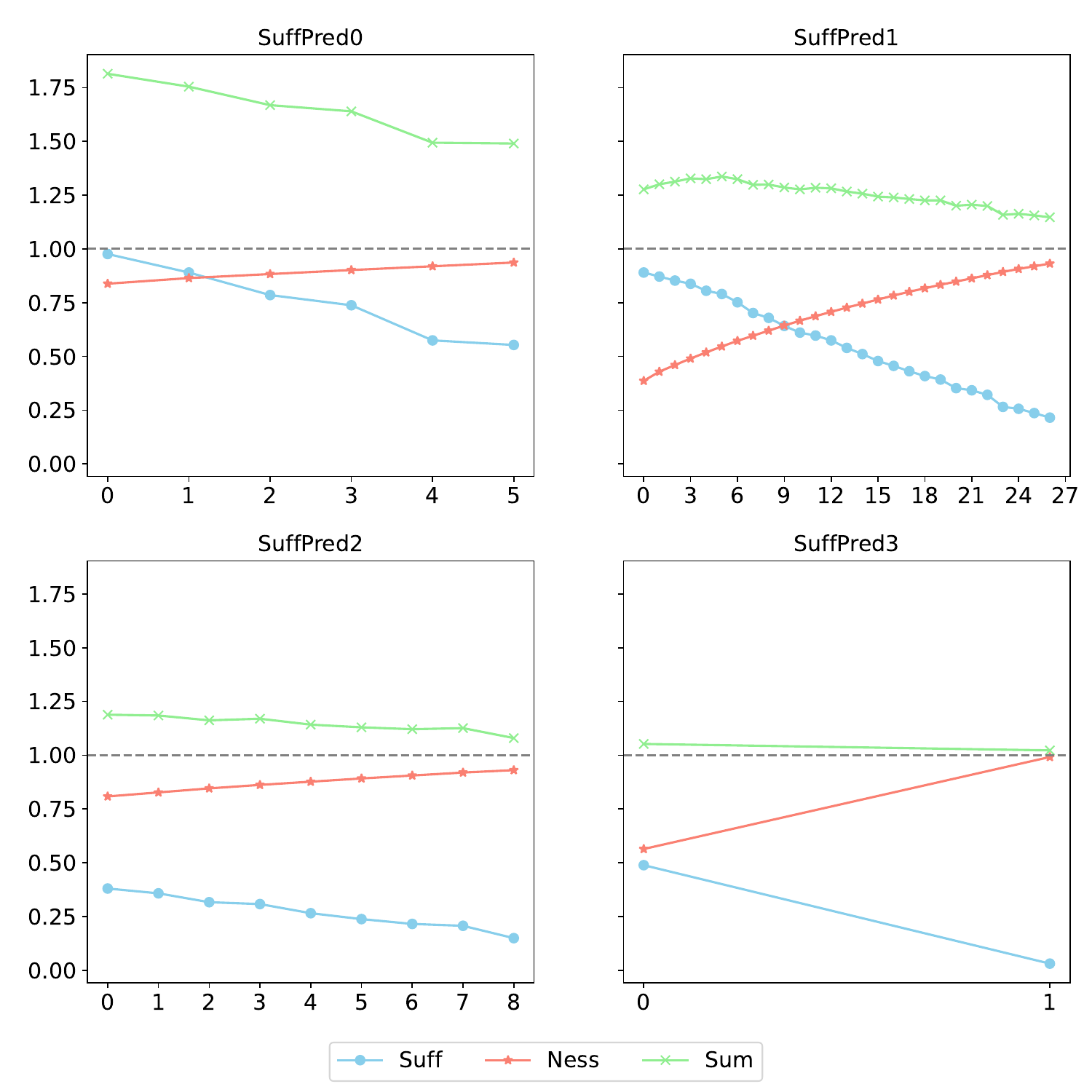}
\caption{\textbf{Sufficient Predicate Scores}. The line charts show the scores of the clusters at each step in sufficiency predicate invention for the game \textit{Getout}. Each chart displays the sufficiency score (blue dot line), necessity score (red asterisks * line) and their sum (green crosses x line) at each step of cluster reduction. From step to step, the necessity scores decrease while sufficiency scores increase. The process stops when the necessity (red line) is above the threshold (0.9 in these figures.)}
\label{fig:suff_pred_scores}
\end{center}
\end{figure}

\textbf{Sufficiency Predicate Invention.} The sufficiency predicates are invented by taking the disjunction of the rules and removing the clauses with the least contribution to the sufficiency score one by one. This process aims to improve the sufficiency score of the resulting predicate while maintaining a reasonable necessity score. 
Figure~\ref{fig:suff_pred_scores} illustrates the score changes at each step as clauses are removed. Initially, the clusters have high necessity but low sufficiency, because the reference ranges are combined. By excluding the clause that contributes the least to sufficiency, the sufficiency score increases step by step. 

Some clusters remain their necessity scores even as the steps progress (such as SuffPred0), whereas some clusters attain their necessity scores close to 0 (SuffPred3). Clusters that fail to maintain a proper necessity score are pruned.

\begin{table}[th!]
\centering
\begin{tabular}{lrr}
\toprule
                & High Necessity & Low Necessity \\
\midrule
High Sufficiency       & \CMARK & \XMARK \\
Low Sufficiency        & \CMARK & \CMARK/\XMARK \\
\bottomrule
\end{tabular}
\caption{Pruning necessity predicates based on their scores.}
\label{tab:strategy_to_keep_predicate}
\end{table}

\textbf{Predicate Selection Strategy.}
Based on the evaluation results, the predicates kept for further use are shown in Table~\ref{tab:strategy_to_keep_predicate}.
Predicates with high necessity and high sufficiency are common in positive states $\mathcal{S}_a^+$ of action $a$ and relatively rare in negative states $\mathcal{S}_a^-$, indicating their significant role in action decision-making. 
Predicate with high necessity, but low sufficiency appear frequently in both positive and negative states and are retained for potential usage in sufficiency invention. 
High sufficiency and low necessity predicates are rare in both state types, while low sufficiency and low necessity predicates, which are often true in negative states but not positive states, can be considered for future work involving negated reasoning.

\paragraph{Example for invented predicates.}
 The following list shows an example of an invented sufficiency predicate for the game \textit{Getout},
\begin{lstlisting}[frame=single,caption={An invented sufficiency predicate for the game Getout}]
InvP1(X):-Dist_[0.04,0.05)(enemy,player,X).
InvP1(X):-Dist_[0.05,0.06)(enemy,player,X).
InvP1(X):-Dist_[0.06,0.07)(enemy,player,X).
InvP1(X):-Dist_[0.07,0.08)(enemy,player,X).
\end{lstlisting}
which can be interpreted as \textit{if the distance between the enemy and the player is in range of 0.04 to 0.08}.
The predicate $\texttt{InvP1}$ evaluates a safe jumping distance with the enemy. 
The corresponding rules have been searched using this predicate for take the action $\texttt{Jump}$:
\begin{lstlisting}[frame=single,caption={Rules using predicate InvP1}]
Jump(X):-InvP1(X),NotExist(key,X),
Dir_[0,4)(enemy,player,X).
Jump(X):-InvP1(X),Dir_[4,8)(enemy,player,X).
Jump(X):-InvP1(X),NotExist(key,X),
Dir_[184,188)(door,player,X).
Jump(X):-InvP1(X),Dir_[184,188)(door,player,
X),Dir_[184,188)(enemy,player,X).
\end{lstlisting}

\begin{figure*}[t]
\begin{center}
\includegraphics[width=0.9\textwidth]{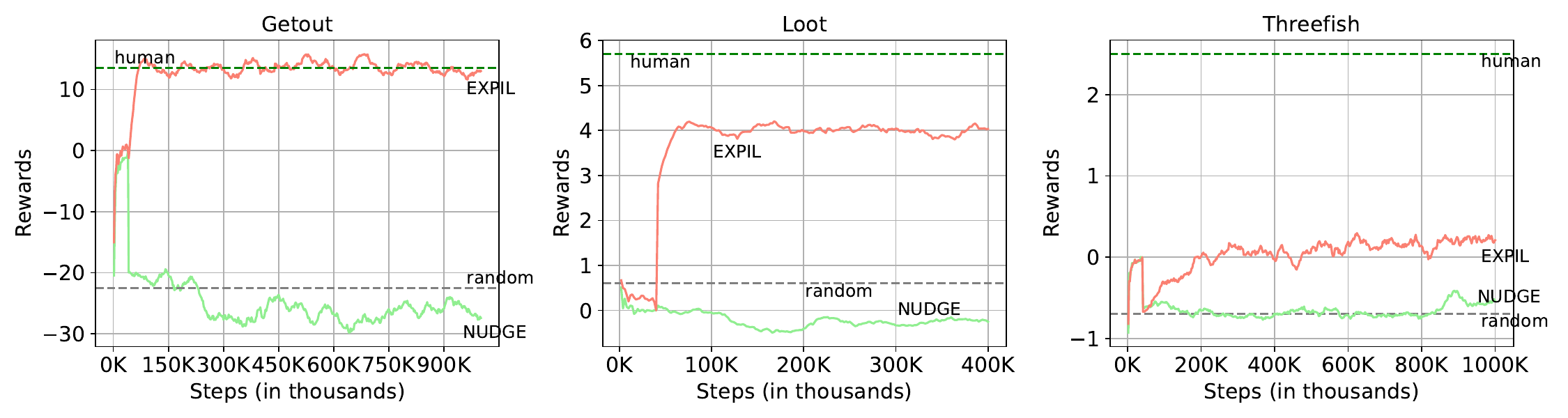}
\caption{
\textbf{The average reward during the training}. The rewards are tracked and smoothed over the last $40,000$ steps to provide a clear trend of performance improvement over time. The red line represents the reward of EXPIL, the green line represents the reward of NUDGE. The gray dash line represents the average reward of a random model, and the green dash line represents the average reward of a human player. 
}
\label{fig:game_rewards}
\end{center}
\end{figure*}
 \begin{table*}[th!]
\centering
\begin{tabular}{l|rrrrrrrr|rrrrr}
\toprule
Game           & $|\mathcal{A}|$ & M. & \#Obj. &  \#N.P. & \#S.P.  & \#R. & \#Dir. & \#Dist. & \textbf{Rand.} & \textbf{Hu.} & \textbf{PPO} & \textbf{NUDGE} & \textbf{EXPIL} \\
\midrule
Getout      & 3 & 1.3\% & 4 & 10 & 1 & 27 & 90 & 100 &    -22.5 & 13.5 & 3.8 & -26.1 & \textbf{13.7}\\ 
Loot        & 4 & 7.3\% & 5 & 32 & 6 & 101 & 8 & / & 0.6 & 5.7 & \textbf{4.5} & -0.2 &  4.3\\
Threefish   & 5 & 14.8\% &  3 & 20 & 5 & 59 & 10 & / &    -0.7 & 2.5 & -0.4 & -0.5 & \textbf{0.4}\\

\bottomrule
\end{tabular}
\caption{\textbf{Model Performance}. 
$|\mathcal{A}|$ denotes the number of actions of each game. 
$M.$ denotes the average relative magnitude of objects in the game map.
$\#O$ denotes the number of objects of each game.
$\#$N.P. denotes the number of necessity predicates used in the policy clauses. 
$\#$S.P. denotes the number of sufficiency predicates used in the policy clauses. 
$\#R.$ denotes the number of reasoned rules. 
$\#$Dir. denotes the number of reference ranges for direction. 
$\#$Dist. denotes the number of reference ranges for distance. 
\textbf{Rand.} shows the average score of a model taking random actions. 
\textbf{Hu.} shows the average score of human players. Higher scores indicate better performance.
}
\label{tab:model_performance}
\end{table*}
Figure~\ref{fig:game_rewards} shows the average game rewards during the weight learning on three logic games. EXPIL achieves similar score as human player on \textit{Getout}, a distinguishable higher score compared to NUDGE player on \textit{Loot} and \textit{Threefish}.
Table~\ref{tab:model_performance} shows the performance of EXPIL in detail.
Our results shows that the EXPIL can invent efficient predicates using predefined physical concepts as background knowledge and further reason rules that achieve high rewards in various logic games.
\textit{Getout} utilizes a largser number of reference range in both direction and distance since the magnitude of objects in this game are relatively small whereas the objects in other games have larger scale (1.3\% in Getout compare to 14.8\% in Threefish). It also invents less predicates and rules since the action space is smaller. \textit{Loot} has reasoned the most rules over three games since it has more objects and 2D map. \textit{Threefish} has the largest action space, but with a relatively simpler objective and largest average object magnitude, it remains fewer predicates and rules. We compared our model against a random agent, human player, and a no predicate invention NeSy-RL model. The average reward scores of each game are shown in the right half of Table~\ref{tab:model_performance}. 





\paragraph{Discussions and Limitations}
Although the reasoned rules are fully explainable, rule selection in states with multiple valid rules is based on neurally learned weights. For example, if both \textit{jump because of enemy} and \textit{right because of key} are valid, the agent chooses based on the weights of these rules. A more explainable agent should logically explain why it selects one action over another. Such strategy reasoning requires more background knowledge and causal reasoning. This is a potential future direction. 
Besides, our focus has been on the concepts of \textit{direction} and \textit{distance}. Exploring other physical concepts like temperature, time and weight for parameter-based predicate generation is also an interesting future work.
\newpage
\section{Related Work}
We revisit relevant studies of EXPIL.
\textbf{Predicate Invention} Inductive Logic Programming (ILP)~\cite{Muggleton91,Muggleton95,Nienhuys97,Cropper20} has emerged at the intersection of machine learning and logic programming. ILP learns generalized logic rules given positive and negative examples using background knowledge and language biases.  
Predicate invention (PI) has been a long-standing problem for ILP and many methods have been  developed~\cite{Stahl93PI,aAthakravi12PI,Cropper2019LearningHL,Hocquette20ijcaiPI,KramerAustrian2007PI,Cropper_Morel_Muggleton_2020aaaiPI,Cropper21PI}, and extended to the statistical ILP systems~\cite{Kok05structure,Kok2007StatisticalPI}.
Recently, differentiable ILP frameworks have been developed to integrate DNNs with logic reasoning~\cite{Evans18,shindo21}, and applied to complex visual scenes~\cite{Shindo2023alphailp,shindo23neumann}.
NeSy-$\pi$~\cite{sha24} integrates  PI with the differentiable ILP systems.
EXPIL is the first PI system on neuro-symbolic RL agents.
\textbf{Neuro-Symbolic RL.}
Relational RL~\cite{Dzeroski01RelationalRL,Kersting04Bellman,KerstingD08PolicyGradient,Lang12RelationalRLModel} has been developed to tackle RL tasks in relational domains. Relational RL frameworks incorporate logical representations and use probabilistic reasoning. In contrast, EXPIL uses differentiable logic programming.
Symbolic programs within RL have been investigated, \eg program guided agent~\cite{sun2020program}, program synthesis~\cite{Zhu19Synthesis}, PIRL~\cite{verma18programmatically}, SDRL~\cite{Lyu19aaai}, 
deep symbolic policy~\cite{Landajuela21icml}, and DiffSES~\cite{zheng2022symbolic}. 
These approaches use domain-specific languages or propositional logic and address either the interpretability.
NUDGE~\cite{DelfosseSDK23nudge} effectively performs a neural-guided search for differentiable logic-based policies to solve complex relational environments, achieving both interpretability and explainability.
EXPIL extends NUDGE, integrating a predicate invention component.

\section{Conclusion}
In this paper, we have proposed EXPIL, a neuro-symbolic framework capable of discovering new concepts while learning to solve RL tasks.
We have introduced two metrics -- necessity and sufficiency -- for EXPIL to invent predicates efficiently from replay buffers recorded by pretrained agents.
By using the invented predicates, EXPIL achieves high-quality logic-based policies with less background knowledge than conventional approaches, making it applicable to various domains.
In our experiments, across three challenging environments where agents need to reason about objects and their relations, 
EXPIL outperforms neural and state-of-the-art neuro-symbolic baselines with zero background knowledge.
We believe that EXPIL would be a basis for intelligent agents that can reason logically and learn with fewer priors, overcoming the bottlenecks of the current neural and neuro-symbolic approaches.





\bibliographystyle{kr}
\bibliography{paper}

\end{document}